\documentclass[11pt]{article}
\usepackage[final]{acl}

\usepackage[T1]{fontenc}
\usepackage[utf8]{inputenc}
\usepackage{newtxtext,newtxmath}
\usepackage[english]{babel}

\usepackage{graphicx}
\usepackage{tabularx} 
\usepackage{booktabs}
\usepackage{multirow}
\usepackage{geometry}
\usepackage[table]{xcolor}

\definecolor{lightred}{rgb}{1.0, 0.4, 0.4}
\definecolor{lightgreen}{rgb}{0.4, 0.7, 0.4}
\definecolor{lightgray}{rgb}{0.5, 0.5, 0.5}

\newcommand{\neutdelta}[1]{\textcolor{lightgray}{(#1)}}
\newcommand{\posdelta}[1]{\textcolor{lightgreen}{(#1)}}
\newcommand{\negdelta}[1]{\textcolor{lightred}{(#1)}}

\title{ALARB: An Arabic Legal Argument Reasoning Benchmark}

\author{
 \textbf{Harethah Abu Shairah\textsuperscript{1}}, 
 \textbf{Somayah AlHarbi\textsuperscript{2}}, 
 \textbf{Abdulaziz AlHussein\textsuperscript{2}}, 
 \textbf{Sameer Alsabea\textsuperscript{1}}, 
 \\
  \textbf{Omar Shaqaqi\textsuperscript{1}}, 
 \textbf{Hebah AlShamlan\textsuperscript{2}}, 
 \textbf{Omar Knio\textsuperscript{1}}, 
 \textbf{George Turkiyyah\textsuperscript{1}}
\\
\\
 \textsuperscript{1}King Abdullah University of Science and Technology (KAUST), 
 \textsuperscript{2}THIQAH
 }

\definecolor{myblue}{RGB}{0, 90, 200}

\begin{document}

\twocolumn[{%
\renewcommand\twocolumn[1][]{#1}%
\maketitle
\includegraphics[width=\textwidth]{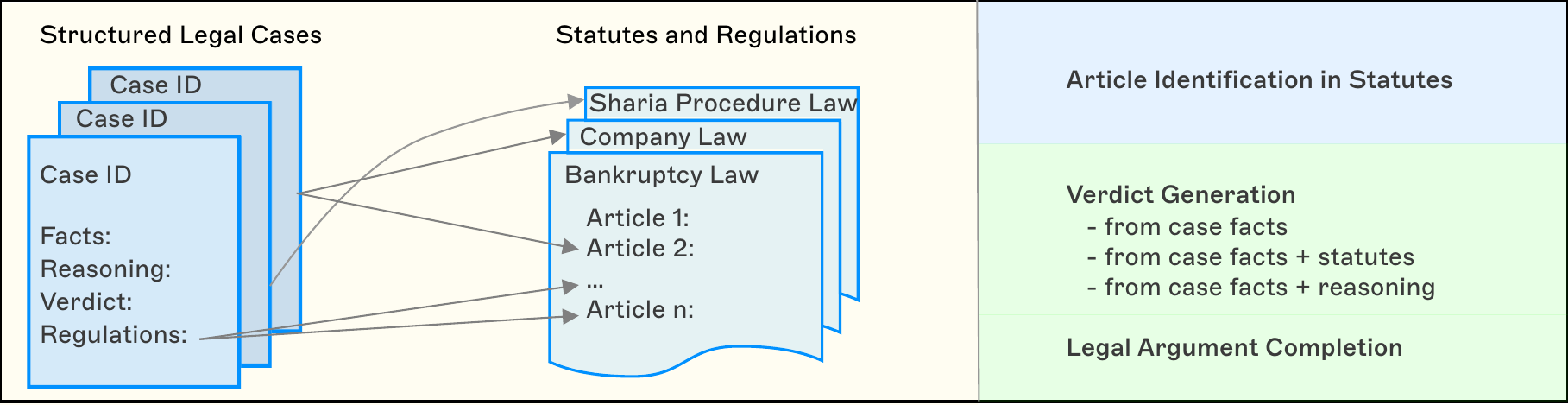}
% \vspace{-2em}
% \captionof{figure}{\small \textbf{ALARB:} The benchmark is constructed from a dataset of structured legal cases. Each case lists the facts presented by the plaintiff and defendant, and an explicit step-by-step chain of the argument reasoning of the court leading to a verdict. Cases are linked to individual articles of applicable statutes and regulations. A set of legal reasoning tasks leverages the data.\vspace{16pt}}
\captionof{figure}{\small ALARB includes a dataset of structured legal cases. Each case lists the facts presented by the plaintiff and defendant, and an explicit step-by-step chain of the argument reasoning of the court leading to a verdict. Cases are linked to individual articles of applicable statutes and regulations. A set of legal reasoning tasks leverages the data. ALARB is available \href{https://huggingface.co/datasets/HarethahMo/ALARB}{\color{myblue}{\textbf{here}}}.\vspace{18pt}}
\label{fig:teaser}
}]

\begin{abstract}
We introduce ALARB, a dataset and suite of tasks designed to evaluate the reasoning capabilities of large language models (LLMs) within the Arabic legal domain. While existing Arabic benchmarks cover some knowledge-intensive tasks such as retrieval and understanding, substantial datasets focusing specifically on multistep reasoning for Arabic LLMs, especially in open-ended contexts, are lacking. The dataset comprises over 13K commercial court cases from Saudi Arabia, with each case including the facts presented, the reasoning of the court, the verdict, as well as the cited clauses extracted from the regulatory documents. We define a set of challenging tasks leveraging this dataset and reflecting the complexity of real-world legal reasoning, including verdict prediction, completion of reasoning chains in multistep legal arguments, and identification of relevant regulations based on case facts. We benchmark a representative selection of current open and closed Arabic LLMs on these tasks and demonstrate the dataset’s utility for instruction tuning. Notably, we show that instruction-tuning a modest 12B parameter model using ALARB significantly enhances its performance in verdict prediction and Arabic verdict generation, reaching a level comparable to that of GPT-4o.
\end{abstract}

% We can insert Arabic \textarabic{مرحبا} inside the otherwise English text. 

\section{Introduction}

The Arabic capabilities of LLMs have been rapidly improving, and many recent models, both closed and open, now demonstrate remarkable fluency and linguistic quality in their generated outputs. This enhanced performance facilitates the development of practical support systems in various knowledge-intensive domains. It also underscores the importance of developing targeted, native Arabic benchmarks to thoroughly evaluate these models in scenarios requiring complex, multistep reasoning.

In English, a variety of benchmarks exist for evaluating the capabilities of emerging LLMs. Several influential benchmarks, such as \cite{wang18_glue, hendrycks21_mmlu}, have significantly shaped the development of earlier models. As these benchmarks quickly become saturated by rapidly improving models—GPT-4.1, for instance, achieves more than 90\% accuracy on MMLU—new benchmarks continue to emerge, offering fresh evaluation challenges \cite{zhong24_agieval, phan2025humanitysexam, guha23_legalbench}. Notably, tasks requiring multistep reasoning have become an essential focus in recent benchmarks, reflecting the capabilities of current-generation LLMs to plan and execute sequences of reasoning steps prior to generating their outputs.

In contract, there is comparatively a dearth of benchmarks to evaluate the emerging generative abilities of Arabic LLMs, and  many existing evaluation and benchmarking resources are in fact translated from English. While in some domains, translations from English or other languages may be quite reasonable, there are others in which LLMs are expected to reason in contexts where social and cultural norms are relevant factors and where translated datasets may suffer from unintended omissions or systematic bias. In order to address this gap, benchmarks that include reasoning tasks in native Arabic contexts are needed. 

The Arabic legal domain provides an ideal setting for benchmarking Arabic LLMs, particularly in open-ended scenarios representative of real-world complexity. 
Legal reasoning involves structured argumentation and contextual sensitivity, and requires flexible inference to handle uncertainties and plausible interpretations that do not exist in mathematical reasoning and inference tasks in closed systems.
Additionally, legal tasks often involve linguistic complexity, nuanced text interpretation, and adherence to formal conventions, further testing Arabic comprehension and generation skills. 
Finally, Arabic remains notably absent from influential multilingual legal datasets \cite{niklaus24_multilegalpile}, underscoring the importance of developing specialized Arabic legal datasets.

Towards this end, we introduce ALARB, a dataset specifically designed to support the multistep reasoning tasks needed for following legal arguments and predicting verdicts. The dataset is derived from original Arabic judicial sources of cases that appeared in front of commercial courts in Saudi Arabia in recent years.

Our contributions can be summarized as follows: 
\begin{itemize}
\vspace*{-6pt}
	\setlength{\itemsep}{-3pt}
	\item We present a 13K+ structured legal cases dataset to support legal argument reasoning, along with their governing statutes.
	
	\item We introduce a set of tasks involving this dataset, including identifying applicable articles from case facts and variants of verdict generation.  
		
	\item We evaluate the performance of the leading open Arabic models on these tasks, and show that the dataset can be used to finetune a 12B model to result in performance that rivals that of GPT-4o.  
\end{itemize}

% Putting this figure here to force latex to put it on page 3
\begin{figure*}
    \centering
    \includegraphics[width=\linewidth]{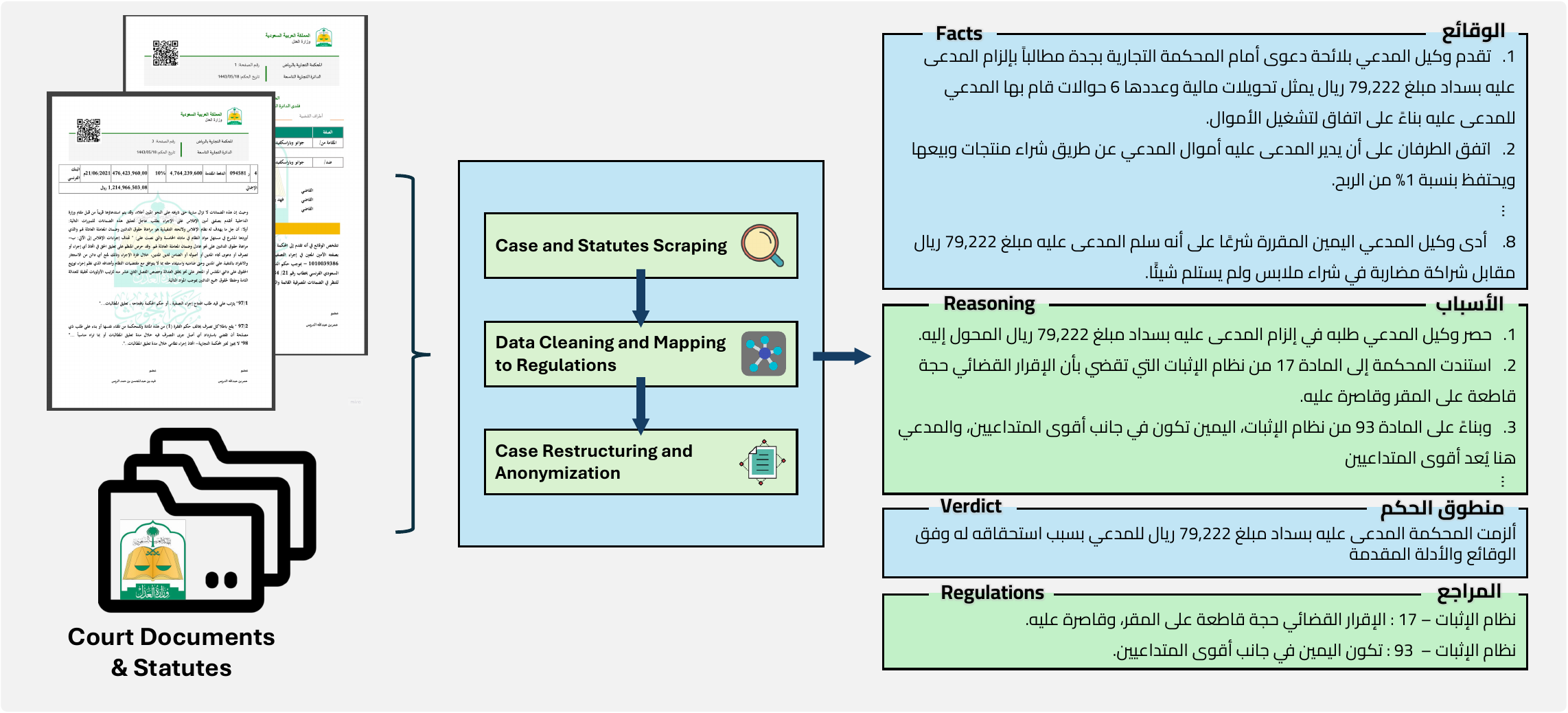}
    \caption{\textbf{Data Preparation Workflow.} \vspace{-6pt}}
    \label{fig:data_processing}
\end{figure*}

\section{Related Work}
\subsection{Arabic LLM benchmarks}
Early benchmarks of Arabic language models largely focused on linguistic-level text classification tasks \cite{antoun20_arabert, abdul-mageed21_dialex} consistent with the limited capabilities of models at the time. Despite interest in evaluating deeper linguistic proficiency \cite{kwon23_beyond,sibaee25_guidelines}, recent benchmarks have shifted towards more knowledge intensive and reasoning tasks to accompany the rising capabilities of current generation Arabic LLMs. In this category of Arabic LLMs, we include both Arabic-centric models \cite{jais23,huang24_acegpt}---models whose training data is mostly focussed on Arabic and English, as well as the multilingual models such as \cite{gemma3, gpt4o, qwen3} that include Arabic among dozens of supported languages. 

Among popular benchmarks for Arabic LLMs, we mention AlGhafa \cite{almazrouei23_alghafa} and ArabicMMLU \cite{koto24_arabicmmlu} that have curated multiple choice questions (MCQs) spanning a variety of general knowledge questions.
% in subject areas generally covered in school curricula in the Arab world. 
The performance of Arabic models on these and other benchmarks are tracked in public leaderboards including the Open Arabic LLM Leaderboard \cite{OALL24} and BALSAM \cite{balsam24}. There has also been interest in benchmarking Arabic LLM models for cultural alignment \cite{qian24_cameleval,mousi25_aradice}.

There is however a need for the evaluation of emerging Arabic LLMs on more challenging tasks that require the generation of conclusions and explanations in open-ended and specialized domains. 
% These tasks go beyond retrieving knowledge the model has or simple summarization, requiring multiple steps of reasoning through the generation process to arrive at outputs that are both semantically and stylistically appropriate. 
A task in the domain of poetry understanding and explanation is described in \cite{alghallabi25_poetry}.

\subsection{Legal reasoning benchmarks and tasks}
The legal domain has seen tremendous interest in the use of LLMs in tasks related to legal research and writing tools targeting professionals and the public, motivating the need for benchmarking in this domain. Early benchmarks \cite{chalkidis22_lexglue,hendrycks21_cuad} focussed on classification and recognition tasks in judgement prediction, clause identification, and related tasks. More recent efforts \cite{guha23_legalbench,fei24_lawbench,nigam24_pred,nigam25_legalseg} have substantially expanded the evaluation tasks to include a broader range of legal reasoning tasks, specifically designed to test logical reasoning, judgment prediction, and question-answering abilities of models. In Arabic, a benchmark inspired by LegalBench appeared in \cite{hijazi-etal-2024-arablegaleval}. 

However, these benchmarks have not addressed tasks that require understanding or generating chains of legal arguments in support of a decision, making it questionable how much legal reasoning of models is being evaluated. In fact, legal LLMs are still prone to hallucinations \cite{magesh25} that are partly attributed to the models' inability to reason correctly through the text to arrive at the proper conclusion. Reasoning-focused datasets and tasks are needed to support reliable RAG systems, explainability, and trustworthiness of LLMs in legal domains. \cite{zheng25,chlapanis24_lar} are efforts in this direction.

\section{Dataset}

The ALARB dataset contains legal cases from commercial courts in Saudi Arabia with their applicable statutes. In this section we describe the process of curating this data and its results.

\subsection{Data Curation}

Figure \ref{fig:data_processing} depicts the data preparation workflow.

\paragraph{Case and Statutes Scraping.} Court case descriptions are scraped from the KSA Ministry of Justice (MoJ) website. Each case description includes the facts of the case (arguments presented by the plaintiff and defendant to the court) and the reasoning of the court.  Each is usually a few paragraphs long. The description also includes a verdict that is short and authoritative in tone. Eight statues, along with their implementing regulations, were identified as the governing documents for these cases and were also scraped. Each of these governing documents is organized into articles representing specific provisions. 

\paragraph{Data Cleaning and Mapping to Regulations.} This involved identifying the statutes and regulation documents, as well as the specific articles from them, that are referenced in each case. These articles are not listed separately in the case descriptions but appear in-line in the text describing the reasoning of the court.  In addition, these articles and their statutes are referred to differently in different cases, with inconsistencies in the naming conventions for the same legal document and in the way article numbers appear in the descriptions. This is essentially a named-entity recognition (NER) task and we used an LLM for it.  Our experiments showed that modern LLMs can generally understand the context needed to identify the statute names and article numbers referred to in the text. For additional robustness however, this process was repeated twice using different prompts, and the union of the two different outputs was used to minimize the risk of missing any relevant articles and regulations. 

\begin{table}
    \centering
    \includegraphics[width=1\linewidth]{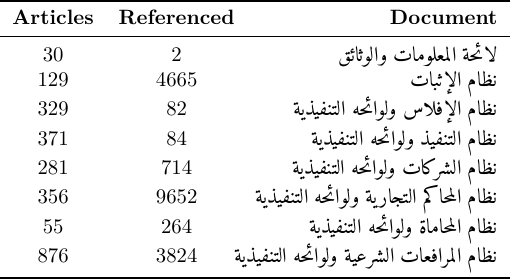}
    \caption{\textbf{Statistics of Referenced Legal Statutes.} 
% The number of articles in each included legal document and how often they were referenced.
}
\label{tab:legal_stats_structured}
\end{table}

\begin{table}
\centering
\small
\resizebox{\columnwidth}{!}{%
\begin{tabular}{lccc|ccc}
\toprule
\multirow{2}{*}{\textbf{Field}} & \multicolumn{3}{c|}{\textbf{Words}} & \multicolumn{3}{c}{\textbf{Steps}} \\
               & Min & Max & Avg & Min & Max & Avg \\
\midrule
Facts        & 31  & 398 & 181 & 3  & 11  & 8 \\
Reasoning      & 18  & 296 & 129 & 1  & 11  & 6 \\
Regulations & 0   & 977  & 186   & 0  & 15  & 3 \\
Verdict     & 5   & 26  & 13  & \multicolumn{3}{c}{N/A} \\
\bottomrule
\end{tabular}
}
\caption{\textbf{Dataset Summary Statistics.} 
% For facts and regulations, steps refers to the number of included facts and regulations in a case.
}
\label{tab:field_stats_structured}
\end{table}

\paragraph{Case Restructuring and Anonymization.}  This involves arranging the facts of a case into a list of individual items, each representing a single fact and generally written in a sentence or two in the text. Similarly, the reasoning was structured as a list of individual steps, each representing a single thought in the reasoning process.  The scraped textual descriptions of the facts and the reasoning also often contained identifiable information about plaintiffs and defendants, which needed to be removed. Prompts were designed to restructure both the facts and reasoning sections into clear steps and to remove irrelevant or sensitive information, and this step was done with an LLM. The quality of the outputs was verified manually on random samples.

Appendix \ref{appendix:dataset} shows an example of the generated representation structured as: a list of individual facts, a sequence of reasoning steps, a court verdict, and keys to full text descriptions of cited articles.

\subsection{Dataset statistics}

Table~\ref{tab:legal_stats_structured} summarizes the data of legal documents included in the dataset. Each entry shows the number of articles contained in the corresponding statute. On average, each article in the statutes has about 47 words. Also shown in the table are the number of times articles from the statute are referenced. In many of the cases, multiple articles from the same statute are referenced.

\begin{figure}
    \centering
    \includegraphics[width=\linewidth]{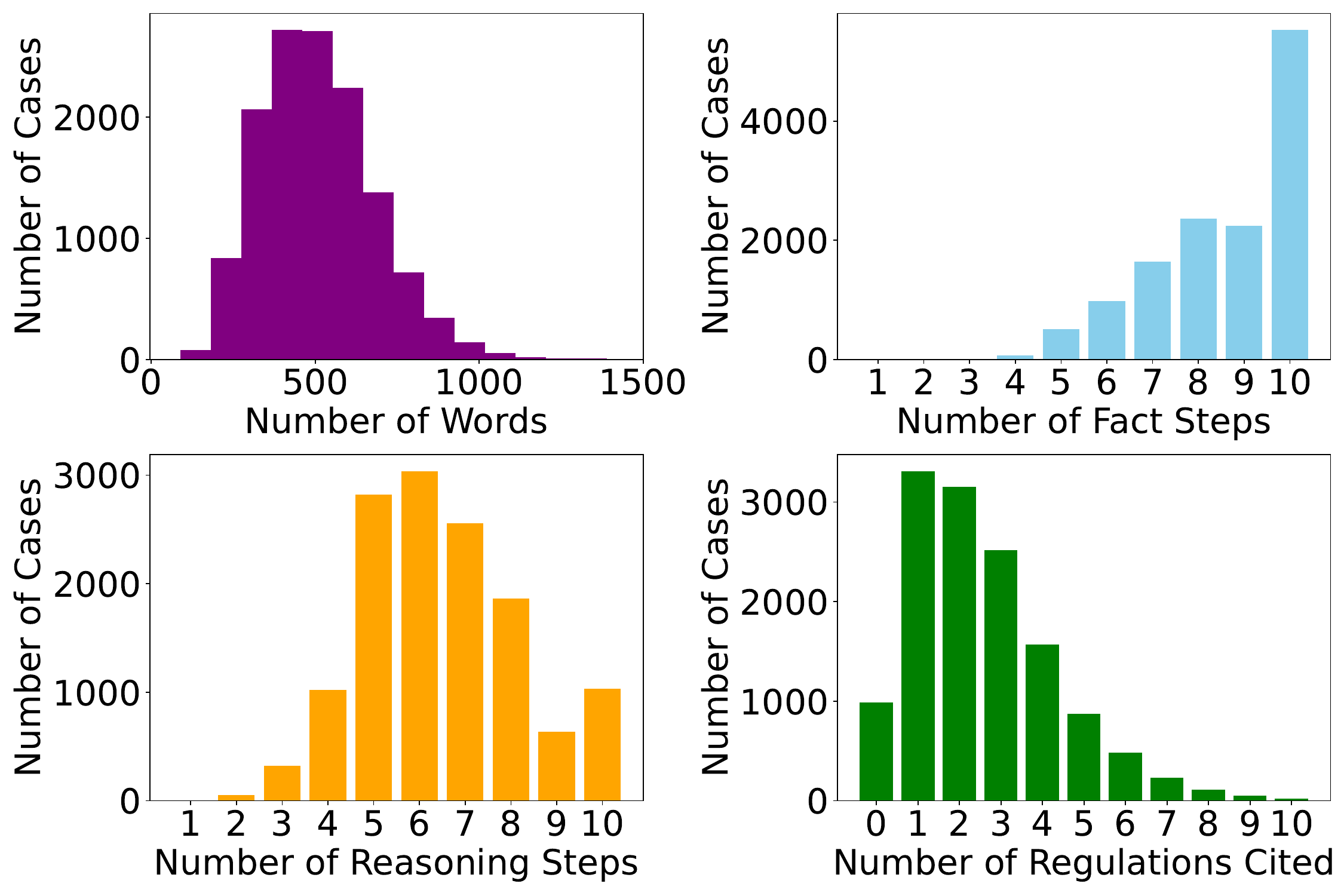}
    \caption{\textbf{Distributions of Words and Steps.}}
    \label{fig:histograms}
\end{figure}

% \vspace*{0pt}

\begin{table}
\centering
\small
\begin{tabular}{ccc}
\toprule
\textbf{For Plaintiff} & \textbf{For Defendant} & \textbf{Court Dismissal}  \\
\midrule
62\% & 5\% & 33\% \\
\bottomrule
\end{tabular}
\caption{\textbf{Case Verdict Breakdown.}}
\label{tab:case_outcomes}
\end{table}

Figure \ref{fig:histograms} shows the composition of the 13,344 legal cases of the dataset. The top left histogram shows their word count distribution, including all text from the list of facts, steps of the reasoning, the verdict, and the referenced articles. There were a few outliers but we had generally chosen cases that are not too lengthy, resulting in the peak of the distribution being around 500 words. The  three other histograms show the distribution of the sizes of the case fact lists, reasoning step lists, and the number of articles explicitly referenced from the statutes. We note that most cases involve about half a dozen discrete reasoning steps and use only a few articles in arriving at the verdict. Table \ref{tab:field_stats_structured} shows additional details of these distributions, with the min, max and average number of words and discrete steps. Table \ref{tab:case_outcomes} shows the verdict distribution of the court rulings, which includes a substantial portion of cases that were deemed not within the court's jurisdiction, with the motivating rationale articulated in the reasoning.

\section{Benchmark Tasks}

ALRAB introduces two main categories of tasks aimed at evaluating a model’s capacity for legal reasoning.

\subsection{Verdict Prediction Tasks}

The first category focuses on verdict generation in different task setups designed to evaluate the models' capacity for legal reasoning with varying amounts of given contextual information. These tasks specifically test how well the model can analyze case details and generate a verdict grounded in the relevant laws and regulations. In each setup, the model is provided with selected information from the case and is expected to produce a legally sound verdict.

\paragraph{Task 1: From Facts Only.} 
In this task, the models are provided with only the factual details of each case. They are expected to analyze these facts to generate a reasoning chain and a verdict solely based on their understanding of the case.

\paragraph{Task 2: From Facts and Relevant Articles.}
In this task, the models receive both the case facts and the specific legal articles that were referenced in the court's reasoning. The objective is to assess the model’s ability to interpret and apply the relevant articles to the facts of a case and produce a reasoned verdict accordingly.

\paragraph{Task 3: From Facts and Court's Reasoning.}
In the setup, the models are given the case facts along with the court’s official reasoning. Based on this combined input, they are tasked with predicting the final verdict. The objective is to evaluate how well they can understand legal arguments in the context of the facts and reach a verdict.   

\paragraph{Task 4: Argument Completion.} 
Tasks 2 and 3 above are two extremes in the spectrum of legal argument reasoning: one provides none of the reasoning of the court and the other provides it all. This task is an intermediate one that provides the models with the first few steps of the reasoning and asks them to complete it and reach a verdict. The task is parameterized by the number of omitted reasoning steps and obviously becomes more difficult as this number increases.

\subsection{Article Identification Tasks}

The second category of tasks is designed to evaluate the models' ability to identify and recognize the appropriate relevant articles in statutes based solely on their understanding of the case facts. To this end, we initially attempted to create a retrieval-based approach where, given only the case facts, the model would retrieve the relevant articles from the entire set of statutes and regulations available. We embedded all available regulations using text-embedding-large-3~\cite{openai2024embedding} and employed cosine similarity to retrieve the most relevant articles based on embedded case facts. However, the results were extremely poor
, which led us to simplify our approach and generate two multiple-choice question tasks instead. 

In these MCQ questions, the models are given the complete list of facts from a legal case and asked to choose the most applicable article from a list of four choices: one being an article actually cited in the court's reasoning and three other distractors. The distractors are constructed in two different ways described below, allowing the MCQs to have two levels of difficulty.

\paragraph{Task 1: Articles from the Same Statute}
In this task, the model is presented with three distractors randomly selected from the same statute as the correct answer. This configuration tests the model's ability to distinguish between somewhat related articles within the same statute. Many articles in the same regulatory document use the same exact words and phrases and require that models understand the full context of an article.

\paragraph{Task 2: Semantically Related Articles}  
In this more challenging task, we employ semantic similarity via embeddings to retrieve articles closely related to the correct article. We utilized the \texttt{text-embedding-large-3} model~\cite{openai2024embedding} for generating embeddings and calculated cosine similarity scores across the entire regulation corpus. The three most semantically similar articles serve as distractors. These may originate from different legal regulations rather than being confined to a single regulatory document. This creates a more sophisticated evaluation that tests the model’s deeper understanding of regulatory nuances, semantic relationships, and subtle differences across various legal texts. A sample MCQ is shown in Figure~\ref{fig:semantic_mcq}.

\begin{table*}
\centering
\resizebox{\textwidth}{!}{%
\begin{tabular}{lccc|ccc|ccc}
\toprule
\multirow{2}{*}{\textbf{Model}} & \multicolumn{3}{c|}{\textbf{Facts Only}} & \multicolumn{3}{c|}{\textbf{Facts \& Reasoning}} & \multicolumn{3}{c}{\textbf{Facts \& Regulations}} \\
& Correct & Partial & Incorrect & Correct & Partial & Incorrect & Correct & Partial & Incorrect \\
\midrule
\texttt{AceGPT-v2-32B-Chat} & 28.9 & 34.7 & 35.8 & 41 & 55.1 & 3.9 & 25.1 & 27.8 & 38.3 \\
\texttt{AceGPT-v2-8B-Chat} & 33.4 & 33.4 & 33.1 & 58.4 & 38.8 & 2.7 & 28.9 & 30.3 & 37.3 \\
\texttt{ALLaM-7B-Instruct-preview} & 14.1 & 42.7 & 43.1 & 39 & 56.4 & 4.5 & 17.2 & 44.3 & 38.4 \\
\texttt{aya-expanse-32B} & 32.9 & 33 & 33.9 & \textbf{70.6} & 26.7 & 2.7 & 36.3 & 32 & 31.5 \\
\texttt{aya-expanse-8B} & 25.6 & 38.8 & 35.6 & 61.9 & 34 & 4.1 & 24.6 & 40.7 & 34.6 \\
\texttt{Falcon3-7B-Instruct} & 8.7 & 20.2 & 70.9 & 28.7 & 40.1 & 31.1 & 8.7 & 18.1 & 73.1 \\
\texttt{Gemma-3-12B-it} & 15.8 & 51.8 & 32.4 & 51 & 46.2 & 2.8 & 29.6 & 40.8 & 29.6 \\
\texttt{Gemma-3-4B-it} & 13.3 & 46.2 & 40.3 & 46.9 & 39.2 & 13.9 & 24.5 & 38.5 & 36.9 \\
\texttt{GPT-4o }& \textbf{38.7} & 31.4 & \textbf{29.9} & 65.7 & 31.6 & 2.7 & \textbf{46} & 28.8 & \textbf{25} \\
\texttt{GPT-o4-mini} & 22.9 & 46 & 30.9 & 61.3 & 36.7 & \textbf{2} & 27.6 & 43.8 & 28.5 \\
\texttt{Qwen3-14B} & 31.5 & 36.5 & 31.9 & 64.5 & 31.5 & 4.1 & 44.6 & 28.7 & 26.7 \\
\texttt{Qwen3-8B} & 27.1 & 36.4 & 36.5 & 58.3 & 36.2 & 5.3 & 32.2 & 34.2 & 33.5 \\
\bottomrule
\end{tabular}%
}
\caption{\textbf{Verdict Prediction results:} LLMs Evaluation for Verdict Prediction Across Three Tasks.}
\label{tab:verdict-prediction}
\end{table*}

\section{Results}
For all tasks, we conducted evaluations across a diverse set of models, varying in size, language capability (Arabic-centric and multilingual), and accessibility (open-source and proprietary). The  list of models included in our evaluation is provided in Table \ref{tab:verdict-prediction}. The benchmarks were performed on a subset of \textbf{1,329} legal cases.

\subsection{Verdict Prediction Tasks Results}

For the first category of tasks—\textit{verdict prediction}—the models were provided with detailed prompts outlining both the expected output and the format of the response. In the two setups where the court’s reasoning was not included as part of the input, the models were explicitly instructed to perform reasoning before generating a verdict.

To evaluate the predicted verdicts, we used \texttt{GPT-4o} as an LLM-as-a-judge \cite{zheng23_llmjudge,gu24_llmjudge}. The model was provided with both the predicted and actual verdicts and tasked with assessing their alignment. 
Reliable automatic evaluation of generated verdicts is not a simple task. 
% We used a conservative approach in our evaluation. The LLM-judge implicitly checks, not only which party won the case or whether the case was dismissed, but that the amounts decided are correct, and that the verdict is generated in the style of the court with no extraneous details. We manually checked the strict judging quality on selected examples. 
Verdicts in commercial cases are not binary and generally require the calculation of fines, which must be done accurately. The judging prompt is shown in % Figure \ref{fig:llm-judge-prompt} 
in Appendix \ref{appendix:prompts}. It generates one of three evaluations:  
\begin{itemize}[noitemsep, topsep=0pt]
    \item \textbf{CORRECT}: The predicted verdict fully matches the actual court verdict.
    \item \textbf{INCORRECT}: The predicted verdict does not align with the actual court verdict. It may award incorrect amounts, not recognize jurisdiction, or add unnecessary details. 
    \item \textbf{PARTIALLY CORRECT}: The prediction demonstrates partial alignment but fails to fully match the court’s decision, mostly in minor style and expression. 
\end{itemize}

% \textcolor{red}{Need to explain (defend?) the use of LLM-as-a-judge. More guidance could have perhaps been given about fine amounts?}.  

In the \textbf{facts-only} task, \texttt{GPT-4o} achieved the highest percentage of correct verdicts, while \texttt{Gemma 3-12B} achieved the highest rate of partially correct predictions.

In the \textbf{facts and court reasoning} task, \texttt{aya-expanse-32B} outperformed all models, followed by \texttt{GPT-4o} in the percentage of correct verdicts. Despite being provided with both the case facts and the court’s reasoning, and only required to interpret the reasoning to reach a verdict, fewer than half of the models achieved more than \textbf{60}\% accuracy. This outcome highlights the inherent complexity of correctly interpreting the dense Arabic legal language of the courts. 

In the \textbf{facts and regulations} task, \texttt{GPT-4o} again led in performance, achieving a \textbf{46}\% correct verdict rate, followed closely by \texttt{Qwen3-14B} at \textbf{44.6}\%. Both models also recorded the lowest percentage of incorrect verdicts, suggesting that they successfully reasoned and applied relevant regulations in approximately \textbf{75}\% of cases.

Interestingly, several models, including both versions of \texttt{AceGPT-v2}, \texttt{aya-expanse-8B}, and \texttt{Falcon-7B}, performed worse when provided with the relevant regulations compared to when they received only the facts. This suggests that the presence of large amounts of legal text in the context may have introduced confusion in models with less robust reasoning capabilities.

Both versions of \texttt{Qwen3} were evaluated with \textit{thinking mode} enabled, allowing us to evaluate the effects of additional test-time reasoning. Under this configuration, the models demonstrated strong reasoning capabilities. \texttt{Qwen3-14B} achieves results that closely approach those of \texttt{GPT-4o}, and both \texttt{Qwen3} models consistently outperform \texttt{o4-mini} across most evaluation cases. Specifically, \texttt{Qwen3-14B} surpasses \texttt{o4-mini} in the percentage of correctly predicted verdicts across all three tasks. In the \textit{Facts and Regulations} task, \texttt{Qwen3-14B} achieves a significantly higher rate of fully correct verdicts—\textbf{44.6}\% compared to \texttt{o4-mini}’s \textbf{27.6}\%—indicating nearly double the accuracy. Even the smaller \texttt{Qwen3-8B} model outperforms \texttt{o4-mini} in this task in terms of fully correct predictions.

Results for the \textbf{argument completion task} with given partial reasoning are discussed in Section \ref{sec:partial}, along with the performance of a fine-tuned model. 

% Overall, Qwen3-14B exhibits slightly stronger performance than its 8B counterpart across all tasks.

\subsection{Article Identification Task Results}

For the regulation identification task, we evaluated a subset of models on 1,159 MCQs for each of the two tasks. In the task where all answer choices were drawn from the same regulatory document, all models demonstrated strong performance, with accuracy exceeding \textbf{80}\%.  \texttt{GPT-4o} achieved the highest accuracy in this setup at 90.42\%, followed by \texttt{GPT-4.1 }at 86.71\%. However, the task became significantly more challenging when semantically similar articles ---retrieved using embedding-based similarity--- were used as distractors. In this more difficult scenario, overall accuracy declined substantially, with \texttt{GPT-4.1} achieving the highest score at 77.30\%. 

Overall, models with strong reasoning capabilities consistently performed well across both task categories, demonstrating their robustness in legal understanding, verdict prediction, and regulatory interpretation.

% \texttt{Qwen3-8B} achieved the highest accuracy in this setup at \textbf{84.6}\%. However, the task became significantly more challenging when semantically similar articles—retrieved using embedding-based similarity—were used as distractors. In this more difficult scenario, overall accuracy declined, with \texttt{Qwen3-14B} having the highest score at \textbf{71.3}\%.

% Overall, models with strong reasoning capabilities, such as \texttt{Qwen3}, consistently performed well across both task categories, demonstrating their robustness in legal understanding, verdict prediction, and regulatory interpretation.

\begin{table}
\centering
% \small
\resizebox{\columnwidth}{!}{%
\begin{tabular}{lcc}
\toprule
\multirow{2}{*}{\textbf{Model}} & \multicolumn{2}{c}{\textbf{Article Identification Accuracy}} \\
&Same Regulation & Semantically Retrieved \\
\midrule
\texttt{AceGPT-v2-8B-Chat} & 81.79 & 52.72 \\
\texttt{Gemma-3-12B-it} & 82.63 & 67.47\\
\texttt{Qwen3-14B} & 82.20 & 71.30 \\
\texttt{Qwen3-8B }& 84.60 & 67.90 \\
\texttt{GPT-o4-mini} & 90.07 & 73.59 \\
\texttt{GPT-4.1} & 86.71 & \textbf{77.30} \\
\texttt{GPT-4o} & \textbf{90.42} & 76.79 \\
\bottomrule
\end{tabular}%
}
\caption{\textbf{Article Identification Results. \vspace{-6pt}} 
% LLMs Evaluation for Regulation Identification Across Two Tasks.
}
\label{tab:regulation-identification}
\end{table}

% \begin{figure*}
%     \centering
%     \includegraphics[width=1\linewidth]{assets/sft_process.pdf}
%     \caption{\textbf{SFT Process:} For the fine-tuning experiment, we create three categories of instructions: 1) Verdict prediction. 2) Reasoning and Verdict prediction. 3) Verdict justification.}
%     \label{fig:sft_process}
% \end{figure*}

\section{Additional Experiments}
We explore the utilization of our dataset in three focused scenarios: Supervised Fine-tuning (SFT), completion of part of the court's reasoning to predict the verdict, and comparing English versus Arabic reasoning capabilities.

\subsection{Supervised Fine-tuning}

% \begin{figure}
%     \centering
%     \vspace*{-6pt}
%     \includegraphics[width=1\linewidth]{assets/sft_example.pdf}
%     \caption{\textbf{SFT Training:} Example from the verdict prediction task.\vspace*{-6pt}}
%     \label{fig:sft_data}
% \end{figure}

\begin{figure}
    \centering
    \vspace*{-6pt}
    \includegraphics[width=1\linewidth]{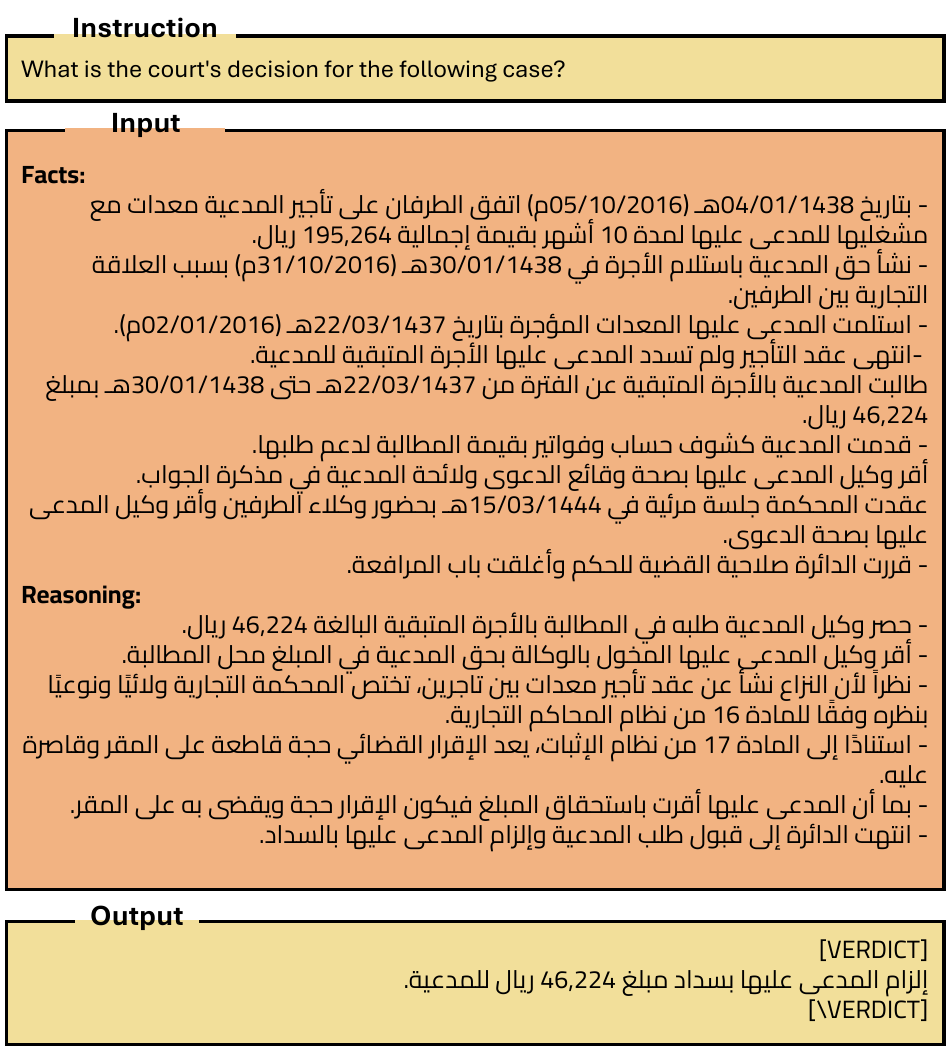}
    \caption{\textbf{SFT Training:} Example from the verdict prediction task. \vspace{-6pt}}
    \label{fig:sft_data}
\end{figure}

A primary application of our dataset is supervised fine-tuning of language models for legal reasoning. To investigate this, we constructed an instruction-tuning dataset derived from the existing cases for SFT and assessed whether fine-tuned models could leverage this dataset to enhance performance on predefined verdict prediction tasks. We initially defined three instruction-based tasks: \textbf{1)} Given legal case facts and applicable regulations, the model generates the reasoning and predicts the verdict. \textbf{2)} Given legal case facts, applicable regulations, and the court's reasoning, the model predicts the verdict. \textbf{3)} Given case facts, applicable regulations, and the final verdict, the model infers the court's reasoning. For task variability, we created multiple instructions per task (details available in Appendix \ref{appendix:sft}). Subsequently, we converted the training portion of our dataset into training samples for instruction-tuning, as illustrated in Figure~\ref{fig:sft_data}. We fine-tuned Google's \texttt{Gemma-3-12B-it} using these instructions and evaluated its performance on our benchmark tasks to measure improvements from fine-tuning.

The model went through full parameter fine-tuning on $12,012$ instruction-output pairs for 4 epochs, with an initial learning rate of $5e^{-6}$ with cosine scheduling, a per device batch size of 2 on 3 A100 GPUs, and 2 gradient accumulation steps.

% \begin{table}[!t]
% \centering
% \resizebox{\columnwidth}{!}{%
% \begin{tabular}{lccc}
% \toprule
% \textbf{Task} &  Correct & Partial & Incorrect \\
% \midrule
% Facts & 39 \posdelta{+9.4} & 32 \neutdelta{-8.8} & 29 \posdelta{-0.6}  \\
% Facts \& Regulations & 39 \posdelta{+9.4} & 32 \neutdelta{-8.8} & 29 \posdelta{-0.6}  \\
% \bottomrule
% \end{tabular}%
% }
% \caption{\textbf{Fine-tuning Impact}: Gemma-3-12B-SFT's performance on the verdict prediction tasks and compared to the base model.}
% \label{tab:english-reason}
% \end{table}

\begin{table*}
\centering
\resizebox{\textwidth}{!}{%
\begin{tabular}{lccc|ccc|ccc}
\toprule
\multirow{2}{*}{\textbf{Model}} & \multicolumn{3}{c|}{\textbf{Facts}} & \multicolumn{3}{c|}{\textbf{Facts \& Reasoning}} & \multicolumn{3}{c}{\textbf{Facts \& Regulations}} \\
& Correct & Partial & Incorrect & Correct & Partial & Incorrect & Correct & Partial & Incorrect \\
\midrule
\texttt{Gemma-3-12B-SFT} & 37.3 \posdelta{+21.5} & 38.6 \neutdelta{-13.2} & 24.1 \posdelta{-8.3} & 65.9 \posdelta{+14.9} & 31 \neutdelta{-15.2} & 3.1 \negdelta{+0.3} & 45.3 \posdelta{+15.7} & 35.7 \neutdelta{-5.1} & 19 \posdelta{-10.6} \\
% Gemma-3-4B-SFT & 31.4  \posdelta{+18.1} & 36.8 \neutdelta{-9.4} & 31.8 \posdelta{-8.5} & 58.8 \posdelta{+11.9}& 32.9 \neutdelta{-6.3} & 8.3 \posdelta{-5.6} & 36 \posdelta{+11.5} & 32.8 \neutdelta{-5.7} & 30.8 \posdelta{-6.1}\\
\bottomrule
\end{tabular}%
}
\caption{\textbf{Fine-tuning Impact}: Gemma-3-12B-SFT's performance on verdict prediction compared to base model. \vspace{-15pt}}
\label{tab:sft-verdict}
\end{table*}

Table \ref{tab:sft-verdict} summarizes the performance of the fine-tuned model on our 1,329 case test set across the three verdict prediction tasks, highlighting performance \textcolor{lightgreen}{gains} and \textcolor{lightred}{drops}. The model demonstrates significant improvements across all three tasks, bringing it up on par with the best models in Table \ref{tab:verdict-prediction}. The biggest improvements are seen in the "Facts" only task, where the model has to work the hardest to reach the correct verdict. These results highlight the effectiveness of these legal cases as a dataset that can be used for instruction tuning for legal reasoning.

\subsection{Partial Reasoning}
\label{sec:partial}

Table~\ref{tab:verdict-prediction} shows a consistent pattern: models consistently exhibit lower rates of incorrect verdict predictions when explicitly provided with court reasoning, compared to when they must infer reasoning independently. To further investigate this behavior, we ran the reasoning completion task testing how the models perform when provided with only a subset of the reasoning steps. Starting with the verdict prediction task involving case facts, applicable regulations, and all $n$ reasoning steps, we progressively removed the final $k \in \{0, 1, \dots, 6\}$ reasoning steps and measured model performance at each stage. 

\begin{figure}
    \centering
    \includegraphics[width=1\linewidth]{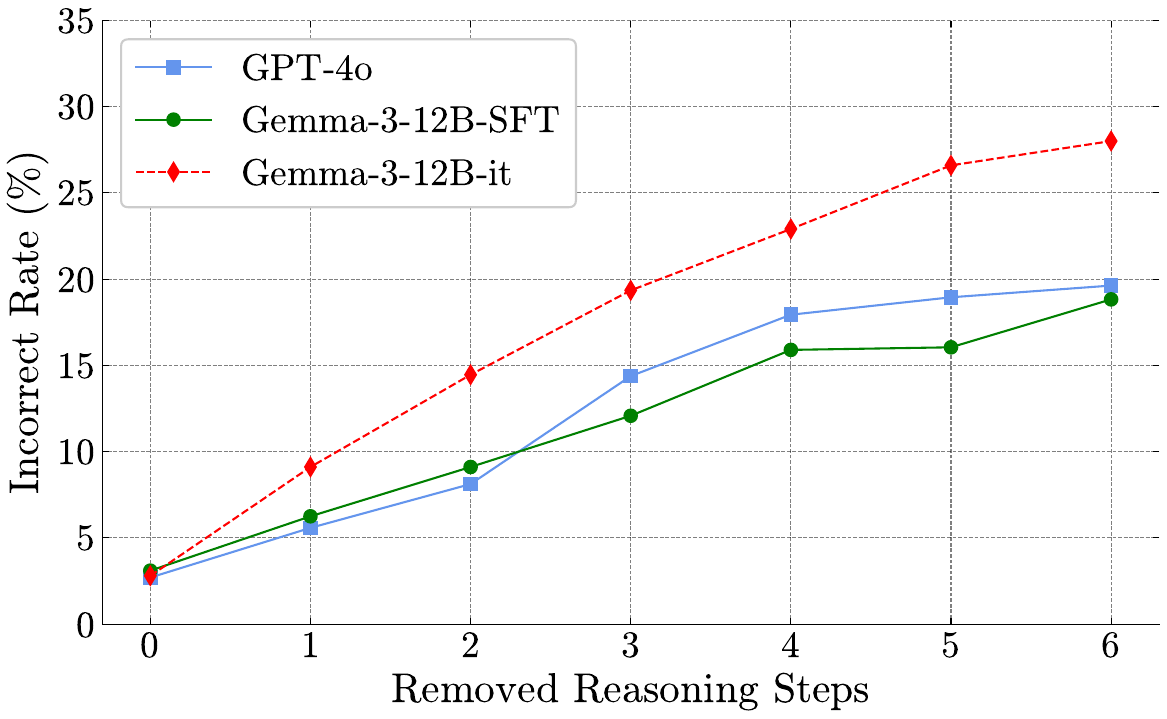}
    \caption{\textbf{Error rate with partial reasoning.\vspace{-6pt}} 
    % The rate of incorrect verdicts when the models are provided with an $n-k$ subset of the court's reasoning steps.
    }
    \label{fig:partial_error}
\end{figure}

Figure~\ref{fig:partial_error} illustrates the increase in error rates as fewer reasoning steps are provided. As anticipated, all models deteriorate in performance when reasoning steps are removed. However, the SFT model demonstrates superior capability at using partial reasoning to reach correct verdicts, surpassing GPT-4o when three or more steps are omitted. 

\subsection{Reasoning In English}

State-of-the-art LLMs are typically trained on extensive multilingual corpora, enabling them to converse and reason across various languages; however, English remains dominant within these datasets. Given that our dataset comprises legal cases exclusively in Arabic, all previously reported results were obtained by explicitly prompting the models to reason and provide verdict predictions in Arabic. We further investigate whether changing the reasoning language from Arabic to English influences model performance. For this experiment, we randomly sampled 100 cases from our test set and used \texttt{GPT-4o} to translate only the verdicts into English, avoiding translation of entire cases due to observed quality degradation in translating legal texts. Using these partially translated cases, we explicitly prompted the models to reason and produce verdict predictions in English for the "Facts \& Regulations" task.

Table~\ref{tab:english-reason} presents the changes in performance for \texttt{GPT-4o} and \texttt{Gemma-3-12B} when reasoning in English. \texttt{GPT-4o} shows minimal variation, with minor performance drops likely attributable to the reduced size of the test sample. On the other hand, \texttt{Gemma-3-12B} exhibits substantial improvement when reasoning in English, significantly increasing its rate of fully correct predictions. This suggests that, despite its multilingual training, \texttt{Gemma-3-12B} benefits greatly from reasoning in English, likely due to stronger linguistic alignment or familiarity. These findings seem to imply that using English reasoning, even for Arabic legal cases, may offer performance advantages for certain multilingual models, as they may be relying on an 
English-centric representation space for their internal reasoning \cite{etxaniz24_multilingual,schut25_do}. 
Further research is needed to reach broader conclusion, however.

\begin{table}
\centering
\resizebox{\columnwidth}{!}{%
\begin{tabular}{lccc}
\toprule
\multirow{2}{*}{\textbf{Model}} & \multicolumn{3}{c}{\textbf{Facts \& Regulations}} \\
& Correct & Partial & Incorrect \\
\midrule
\texttt{Gemma-3-12B-it} & 39 \posdelta{+9.4} & 32 \neutdelta{-8.8} & 29 \posdelta{-0.6}  \\
\texttt{GPT-4o} & 45 \negdelta{-1} & 27 \neutdelta{-1.8} & 28 \negdelta{+3} \\
\bottomrule
\end{tabular}%
}
\caption{\textbf{Reasoning In English:} English reasoning improves Gemma3's performance.\vspace{-6pt}}
\label{tab:english-reason}
\end{table}

\section{Conclusions}
We introduced ALARB, a novel Arabic dataset specifically designed to benchmark legal reasoning capabilities in Arabic LLMs. The dataset features multiple variants of verdict prediction tasks, assessing models’ abilities to comprehend legal linguistic nuances, accurately apply regulations to given cases, and produce legally sound reasoning chains. Our experiments demonstrate that reasoning-oriented models generally perform better on these tasks; however, significant opportunities for improvement remain. Additionally, we validated ALARB’s effectiveness by fine-tuning a 12B-parameter model, resulting in substantial performance gains. For future work, we intend to leverage ALARB in the Reinforcement Learning (RL) post-training of Arabic reasoning models.
% to further enhance their reasoning proficiency.

\section*{Limitations}
While this study contributes to evaluating and improving Arabic LLMs, several limitations must be acknowledged and addressed in future work.

First, the dataset is limited to a particular area of the law, obtained from a single country, and is relatively limited in size. Additional diversity is needed to broaden its capabilities. Texts from some areas besides commercial law are publicly available and may be used. Ministries of Justice in many countries of the Arab world have digitized their documents and these represent valuable resources for expanding and enriching the dataset with different styles of reasoning. 

Evaluation of the LLM-as-a-judge in verdict prediction tasks merits deeper scrutiny. Instead of the ternary classification we used, a finer scale evaluation may be possible, perhaps separating the substance of the verdict from its expression and form. 

When showcasing the effectiveness of the dataset for model finetuning, we used a mid-sized model (Gemma-3-12B-it), primarily for convenience. Larger models need to be investigated to further evaluate its utility. 

The reasoning capabilities of the existing Arabic LLMs warrant deeper examination. Our observations of reasoning traces from open models performing test-time inference are that models often pursue incorrect reasoning paths before self-correcting based on additional information, particularly evident when answering multiple-choice questions or applying an article to a case. More thorough analysis is needed to better understand these reasoning dynamics. 

Finally, an intriguing question remains regarding the underlying reasons behind the models' improved performance when prompted to reason in English, and how general this behavior is. 

\section*{Acknowledgment}
We thank Fairs Hijazi for valuable contributions to the early discussions that helped guide and shape this project.

\section*{Ethics Statement}
Legal matters are inherently sensitive and require careful handling. We have anonymized the generated dataset to remove all identifying information about plaintiffs, defendants, as well as the judges that ruled on the cases included. All contributors to this work are properly recognized, either as co-authors or in the Acknowledgments section. 

% \section*{Acknowledgments}
% We thank Faris Hijazi for insightful discussions about early versions of this work.

\nocite{*}

% \clearpage
\bibliography{references,gt_references}

\clearpage
\onecolumn
\appendix

\section{Sample Case from the Dataset}
\label{appendix:dataset}

\begin{figure}[h!]
    \centering
    \includegraphics[width=0.75\columnwidth]{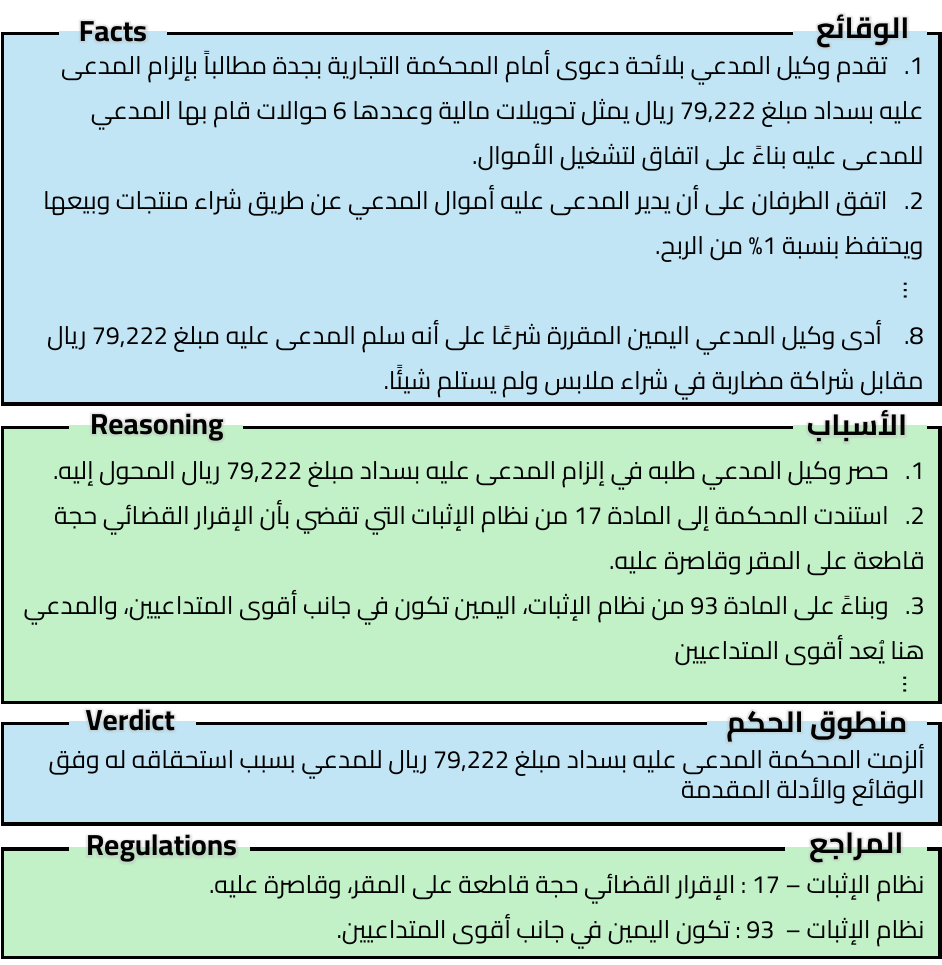}
    \caption{\textbf{Cases Example:} Sample legal case after restructuring.}
    \label{fig:sample_case}
\end{figure}

Figure \ref{fig:sample_case} shows an example of the resulting structured representation of cases.  To support reasoning tasks, each legal case is structured into: (1) a list of individual facts and arguments presented to the court; (2) a sequence of steps articulating the reasoning of the court; (3) the final verdict reflecting the court’s opinion; and (4) the individual articles form the statutes explicitly cited in the case. The cases reference a core set of eight statutes and regulatory documents. Shown in the figure are the (standardized) keys to full text descriptions of statute articles. For convenience, these descriptions have been inserted in the output so every case has the complete reasoning context.  

\clearpage

\section{Prompts for Inference and Evaluation}
\label{appendix:prompts}

\subsection{LLM as a Judge}

\begin{figure}[h!]
\centering
\begin{tabularx}{\textwidth}{X}
\toprule
You are a legal assistant. You will be given a judge's verdict from a legal case in Saudi Arabia, and a prediction of the verdict from another legal assistant. \\
Your task is to evaluate how well the prediction matches the judge's verdict. \\
The evaluation should be based on the content of the verdicts and how well they align with each other. \\
A prediction is correct if it is similar to the judge's verdict and captures the essence of the decision. It does not have to be identical, but it should reflect the same outcome and reasoning. \\
It's acceptable for the prediction to be shorter or more concise than the judge's verdict, or the other way around, as long as the core message is the same. Ignore any noise or irrelevant tokens in the verdicts. \\
Before you output your evaluation, think about how well the prediction matches the judge's verdict. \\
\\
Output one of the following for the evaluation: \\
- \texttt{"CORRECT"} if the prediction matches the judge's verdict.\\
- \texttt{"INCORRECT"} if the prediction does not match the judge's verdict.\\
- \texttt{"PARTIALLY CORRECT"} if the prediction is partially correct but does not fully match the judge's verdict.\\
\\
Follow this format: \\
\texttt{[THINK]} \\
\texttt{"Your reasoning here"} \\
\texttt{[EVALUATION]} \\
\texttt{"Evaluation here (CORRECT, INCORRECT, or PARTIALLY CORRECT)"} \\
\\
Judge's verdict: \\
\texttt{\{judge\_verdict\}} \\
\\
Predicted verdict: \\
\texttt{\{predicted\_verdict\}} \\
\\
Begin! \\
\bottomrule
\end{tabularx}
\label{fig:llm-judge-prompt}
\caption{\textbf{LLM as a Judge Prompt}: The prompt we use for automatic evaluation of verdicts, provide the predicted and court verdicts to the LLM and ask to think before giving an evaluation.}
\end{figure}

\newpage
\subsection{Verdict Prediction}

\begin{figure}[h!]
\centering
\begin{tabularx}{\textwidth}{X}
\toprule
You are a legal assistant specialized in Saudi Arabian law. Your task is to predict the verdict of a legal case from Saudi Arabia. \\
The cases involve trade and finance and commercial laws.\\
\\
You will be given a set of facts from the case, and you MUST provide BOTH:\\
1. A reasoning section analyzing the facts\\
2. A verdict prediction section stating what you think the court will decide\\
\\
The verdict should be based only on the facts provided without personal opinions or biases.\\
Think carefully about the facts and how they relate to the laws in Saudi Arabia.\\
Your verdict and reasoning should be strictly in \texttt{\{language\}}\\
The verdict should be short and direct.\\
\\
Follow the format below:\\
\texttt{[REASONING]}\\
"Your reasoning and analysis here"\\
\texttt{[\textbackslash REASONING]}\\
\\
\texttt{[VERDICT]}\\
"Your verdict here"\\
\texttt{[\textbackslash VERDICT]}\\
\\
Do not output anything else outside these two sections.\\
\\
Here are the facts of the case:\\
\texttt{\{case\_facts\}}\\
\\
Begin!\\
\bottomrule
\end{tabularx}
\label{fig:verdict-facts-prompt}
\caption{\textbf{Prompt for Verdict Prediction from Case Facts.}}
\end{figure}

\newpage

\begin{figure}[h!]
\centering
\begin{tabularx}{\textwidth}{X}
\toprule
You are a legal assistant. Your task is to predict the verdict of a legal case from Saudi Arabia. \\
The cases involve trade and finance and commercial laws.\\
You will be given a set of facts from the case, and the reasoning of court on these facts.\\
You should provide a verdict based on the facts and the reasoning of the court.\\
The verdict is a sentence that summarizes the outcome of the case showing what do you think the court will decide.\\
The verdict should be based on the facts and reasoning provided and should not include any personal opinions or biases.\\
Your verdict should be strictly in \{language\}.\\
Your output should only be a direct and short verdict, do not output anything else.\\
Make sure to label the start and end of the verdict properly.\\
\\
Follow the format below:\\
\texttt{[VERDICT]}\\
"Your verdict here"\\
\texttt{[\textbackslash VERDICT]}\\
\\
Do not output anything else.\\
\\
Here are the facts of the case:\\
\{case\_facts\}\\
\\
Here is the reasoning of the court:\\
\{case\_reasoning\}\\
\\
Begin\\
\bottomrule
\end{tabularx}
\label{fig:verdic-fact-reasoning-prompt}
\caption{\textbf{Prompt for Verdict Prediction from Case Facts and Reasoning of the Court.}}
\end{figure}

\newpage

\begin{figure}[h!]
\centering
\begin{tabularx}{\textwidth}{X}
\toprule
You are a legal assistant. Your task is to predict the verdict of a legal case from Saudi Arabia. \\
The cases involve trade and finance and commercial laws.\\
You will be given a set of facts from the case, and the laws and regulations applicable to this case, and you MUST provide BOTH:\\
  1. A reasoning section analyzing the facts\\
  2. A verdict prediction section stating what you think the court will decide\\
\\
You should provide a verdict based on the facts and the given laws.\\
The verdict is a sentence that summarizes the outcome of the case showing what do you think the court will decide.\\
The verdict should be based on the facts and laws provided and should not include any personal opinions or biases.\\
Your verdict and reasoning should be strictly in {language}.\\
Think about the case facts and how they relate to the given laws.\\
\\
Follow the format below:\\
\texttt{[REASONING]}\\
"Your reasoning and analysis here"\\
\texttt{[\textbackslash REASONING]}\\
\\
\texttt{[VERDICT]}\\
"Your verdict here"\\
\texttt{[\textbackslash VERDICT]}\\
\\
Do not output anything else.\\
\\
Here are the facts of the case:\\
\{case\_facts\}\\
\\
Here are the laws related to this case:\\
\{case\_laws\}\\
\\
Begin!\\
\bottomrule
\end{tabularx}
\label{fig:vereict-facts-laws-prompt}
\caption{\textbf{Prompt for Verdict Prediction from Case Facts and Applicable Regulations.}}
\end{figure}

\clearpage
\section{SFT Instructions}
\label{appendix:sft}
\begin{table}[h!]
\centering
\small
\begin{tabularx}{\textwidth}{lX}
\toprule
\textbf{Task} & \textbf{Instruction} \\
\midrule
\multirow{5}{*}{Verdict Prediction} & What is the court's decision for the following case? \\
& Given the information, how should the court rule, and why? \\
& Based on the facts and reasoning, what is the final verdict of the court? \\
& Analyze the case details and provide the court's verdict. \\
& Given the facts and reasoning, what is the court's decision? \\
\midrule
\multirow{7}{*}{Reasoning \& Verdict Prediction} & Given the following facts and laws, provide the verdict. \\
& Read the facts and applicable laws below, then summarize the court's decision. \\
& Given the case details, generate a summary of the reasoning and the final verdict. \\
& Analyze the following facts and laws, then provide your reasoning and the verdict.\\
& What is the court's decision for the following case? Include reasoning.\\
& After reviewing the facts and applicable laws, explain the court’s reasoning process and final decision.\\
\midrule
\multirow{7}{*}{Verdict Justification (Reasoning)} & Given the facts, laws, and final verdict, explain the legal reasoning of the court step by step. \\
& Analyze the case details and provide a detailed explanation of the court's reasoning leading to the verdict. \\
& Explain the court's reasoning process based on the provided facts, laws, and final verdict.\\
& Given the case facts and laws, summarize the court's reasoning and how it led to the final verdict. \\
\bottomrule
\end{tabularx}
\caption{\textbf{Categories of Instructions for SFT.}}

\end{table}

\clearpage
\section{Sample MCQ from Article Identification Task}
\label{appendix:MCQ}

% \begin{figure*}[h!]
%    \centering
%    \includegraphics[width=\linewidth]{assets/Figure_Example_of_Semantically_Related_Articles_Figure.pdf}
%    \caption{Example showing how overlapping and similar regulations led to failure in semantic retrieval.}
%    \label{fig:semantic_example}
% \end{figure*}
\begin{figure*}[h!]
   \centering
   \hspace*{-5cm}\includegraphics[width=1.6\linewidth]{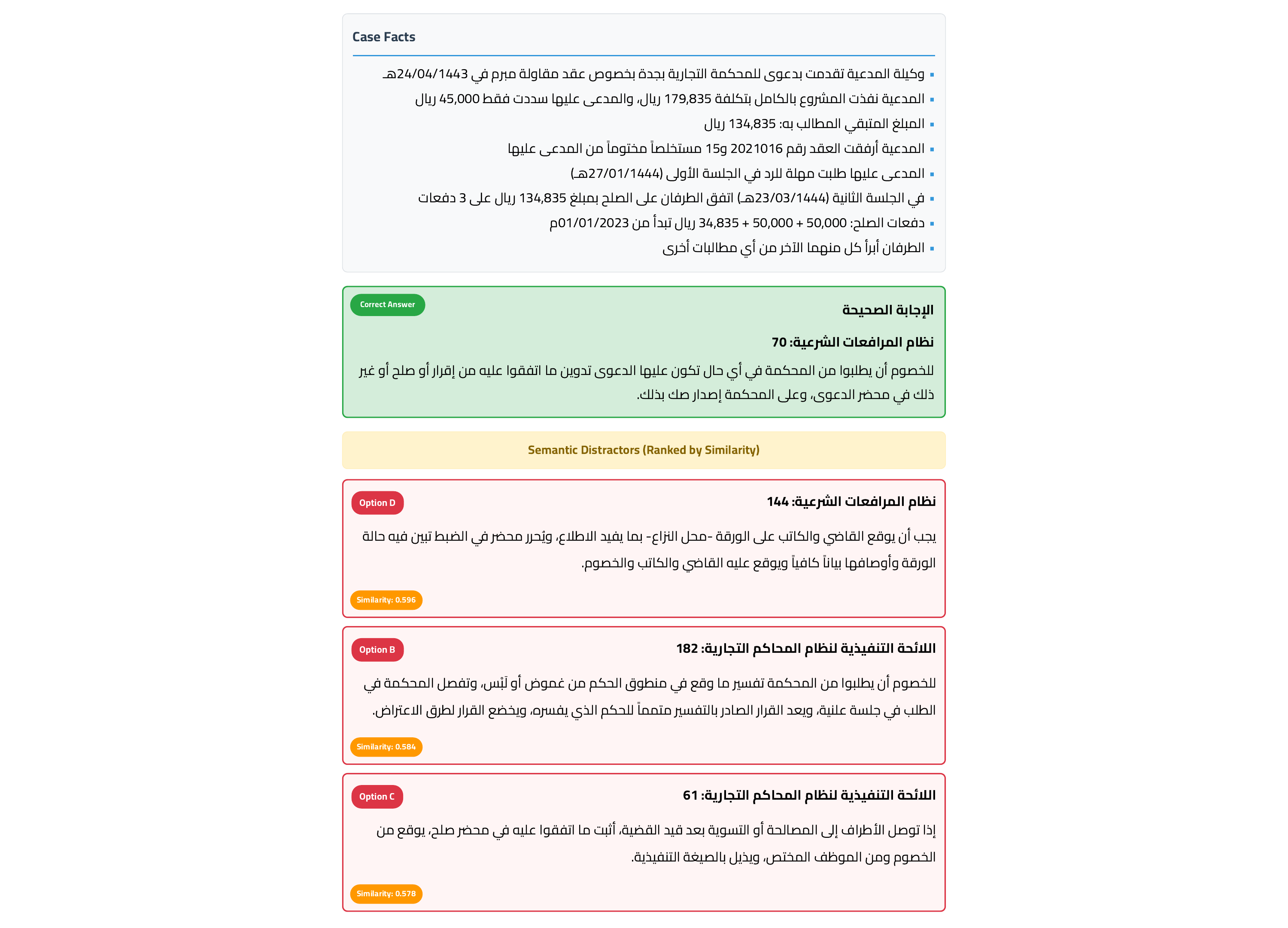}
   \caption{Sample MCQ showing semantically similar distractors}
   \label{fig:semantic_mcq}
\end{figure*}

% \section{Tasks Setup and Evaluation}
% \begin{figure}[H]
%     \centering
%     \includegraphics[width=0.75\linewidth]{assets/llm_judge.pdf}
%     \caption{LLM-as-Judge Prompt}
%     \label{fig:enter-label}
% \end{figure}

% \label{appendix:tasks_eval}

\end{document}